\documentclass[letterpaper,conference]{IEEEtran}
\usepackage{cite}
\usepackage{amsmath,amssymb,amsfonts}
\usepackage{algorithmic}
\usepackage{graphicx}
\usepackage{textcomp}
\usepackage{xcolor}
\usepackage{hyperref}
\begin{document}

\title{Mission Planner for UAV Battery Replacement
}

\author{\IEEEauthorblockN{Zden\v{e}k Bou\v{c}ek and Miroslav Fl\'\i dr}\\
    \IEEEauthorblockA{New Technologies for the Information Society Research Center \& Department of Cybernetics\\
        Faculty of Applied Sciences\\
        University of West Bohemia\\
        Pilsen, Czechia\\
        Email: zboucek@kky.zcu.cz, flidr@kky.zcu.cz}}

\maketitle

\begin{abstract}
The paper presents a mission planner for an autonomous unmanned aerial vehicle (UAV) battery management system. The objective of the system is to plan replacements of the UAV's battery on the static battery management stations. The plan ensures that UAVs have sufficient energy to fulfill their long-term mission, which would otherwise be impossible. 

The paper provides a detailed description of the mission planner and all of its components. The functionality of the planner is successfully demonstrated in simulated multi-UAV multi-station scenarios.
\end{abstract}

\begin{IEEEkeywords}
aerial robotics, battery management, mission planning, UAVs, drones
\end{IEEEkeywords}

Code: \url{https://github.com/zboucek/uav-battery-planner}

\section{Introduction}

The ability to deploy and operate multiple unmanned aerial vehicles (UAVs) simultaneously for extended periods is highly advantageous in a variety of applications, including surveillance, search and rescue, and environmental monitoring \cite{Shakhatreh2019,Abdelkader2021}. However, the management of a swarm of UAVs presents significant challenges, including ensuring safe operations and minimizing operational costs, particularly concerning the number of human operators required. A significant limitation on the endurance of UAVs is their reliance on onboard power sources, which are typically lithium polymer (LiPo) batteries. These batteries typically provide flight times ranging from a few minutes to an hour, depending on the UAV's size and payload.

Various approaches have been proposed to extend the endurance of UAVs, including recharging batteries on the ground \cite{Malyuta2020,Wang2021} or using automated battery replacement stations \cite{Ure2015,Toksoz2011,flytnow}. While recharging allows for continuous operation, it renders the UAV unavailable for missions during charging periods and monopolizes the station for other UAVs. Battery replacement systems offer a more efficient solution but often lack flexibility, supporting only specific UAV models and battery types \cite{Coppola2020}.

As highlighted in \cite{Coppola2020}, extending the flight time of UAV systems remains a fundamental challenge, involving the design of integrated UAV-recharging ecosystems and distributed scheduling for recharging or replacement. The authors note the absence of an automated and distributed recharging method for UAV swarms outside controlled environments. Recent work by Hoang et al. \cite{Hoang2024} explores the potential of leveraging existing power line infrastructure for charging specially equipped UAVs, which could provide an established outdoor charging network.

In our previous endeavors \cite{Droneport2021icr,Severa2022,Blaha2022,Blaha2023}, we introduced the Droneport system, an open-source platform for autonomous battery replacement compatible with various vertical take-off and landing (VTOL) UAVs and battery types. Building on this foundation, the current paper focuses on the software component for intelligent scheduling of battery replacements based on UAV missions and battery statuses.

The proposed scheduling algorithm determines the optimal moments to interrupt user-defined UAV missions for battery replacements, without directly controlling the missions themselves. In contrast to techniques such as Mixed-Integer Linear Programming (MILP) \cite{Song2014} or other optimization methods that plan the overall mission, our approach leverages the well-known A* algorithm \cite{Klancar2017} to efficiently find the optimal times for battery replacements, considering the UAVs' current states and mission progress.

The objective of our research is to develop an automated and intelligent scheduling solution for battery replacements. The objective of this solution is to address the challenges of extending UAV flight time while maintaining mission continuity and minimizing disruptions to user-defined missions. While existing research addresses specific scenarios such as area coverage \cite{Zuo2020,Ghazzai2019}, long-range flights with charging stations \cite{Ghaharikermani2020}, mobile charging stations \cite{Qin2021}, and energy-aware planning \cite{Datsko2024}, our system is designed to provide services that maximize UAV flight time without organizing the missions themselves.

The paper is structured as follows. Section \ref{sec:planner} provides a comprehensive overview of the mission planner, including an in-depth analysis of its fundamental components, underlying principles, and the specific aspects of A* implementation. Section \ref{sec:results} presents the results of two distinct scenarios. The initial scenario in Section \ref{subsec:mission-random} is based on missions where the waypoints (WPs) are randomly generated on an ellipse and each UAV has different parameters. In the second scenario in Section \ref{subsec:mission-park}, software for mission planning was employed to design missions that more closely resembled a realistic operational scenario with the UAV parameters set according to the existing UAV. Finally, the results achieved and insights gained about the mission planner's properties and potential future enhancements to the system are discussed in Section~\ref{sec:conclusion}.

\section{Mission Planner Architecture and Components}
\label{sec:planner}

This section provides a comprehensive description of~the~mission planner. First, the structure of the system is introduced to clarify the basic principles and essential components. Secondly, a method for predicting the battery state based on the UAV's behavior is presented. Thirdly, the~procedures incorporated into the system are described, which reflect the requirements of long-term UAV missions. Finally, the~A*~algorithm itself is described in detail.

\subsection{Virtual Components of System}
\label{subsec:components}

\begin{figure}
    \centering
    \includegraphics[width=\linewidth]{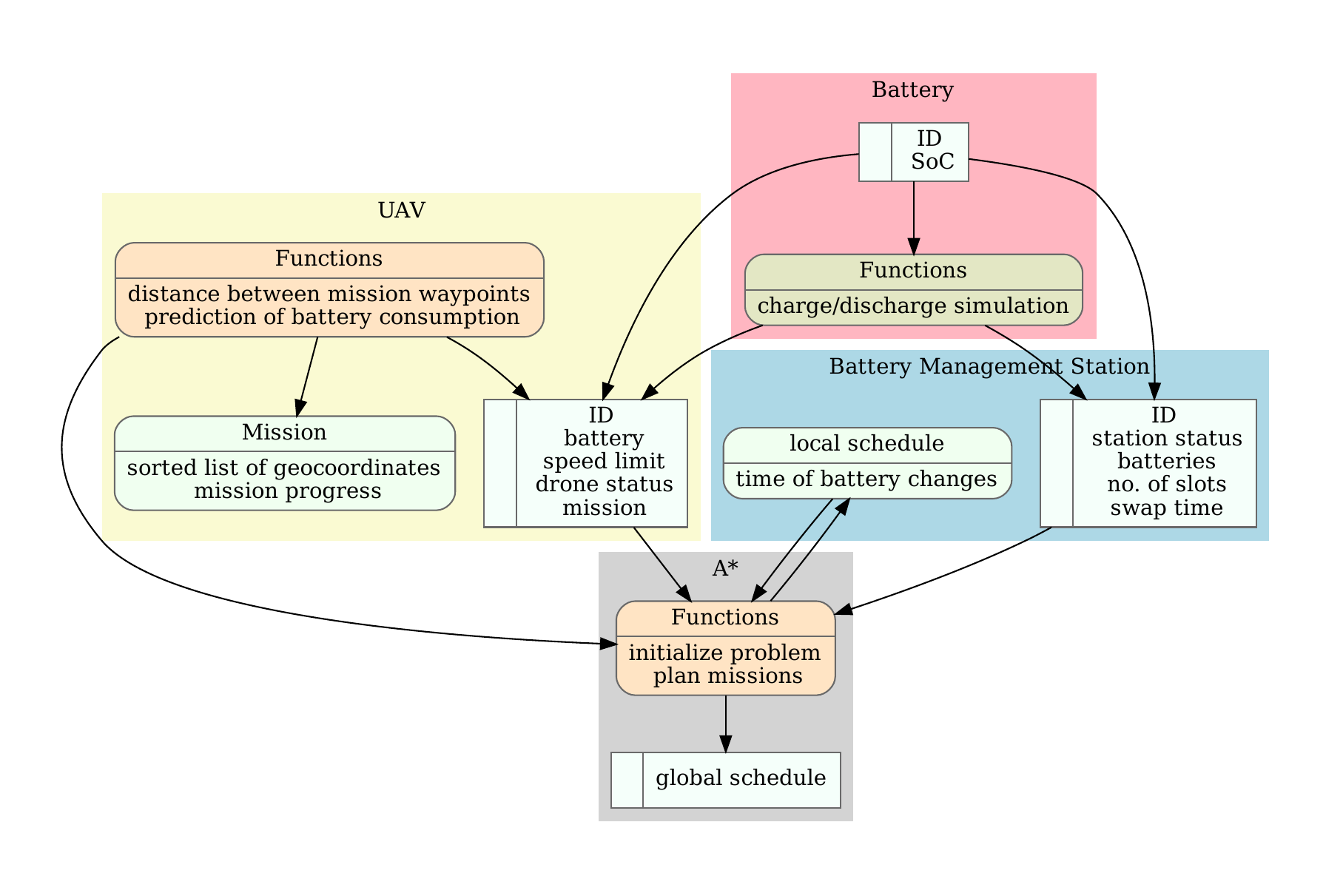}
    \caption{Representation of battery management system in mission planner}
    \label{fig:system-representation}
\end{figure}

The implementation includes objects that represent three key elements of the system: the battery, the UAV, and the battery management station. The structure of the system and the interconnections of its components are depicted in Figure~\ref{fig:system-representation}. The objects facilitate future integration into a larger system and serve as an abstract representation of the physical functional components of the system as a whole.  

This model enables the mission planner to make various predictions about the behavior of the system, which can then be incorporated into algorithms other than A*. Furthermore, this model can be extended in the future, allowing for the addition of new components and improvements to the system.

The UAV object includes methods for calculating distances and simple estimation of battery consumption during a mission. The battery, in turn, includes simple methods to simulate discharging or charging that changes the battery's State of Charge (SoC). The battery management station provides information regarding its operational parameters, including the number of available slots, the status of the installed batteries, and the duration of the battery replacement process. Additionally, the station's local schedule of its occupation is accessible.

Both the UAV and the station contain a variable that stores the current state. For the UAV, this variable indicates whether it is taking off, flying a mission, or waiting for the station to replace a discharged battery with a new one. While they are not currently utilized, these states could potentially be reflected in a real-world application, thereby enhancing the overall effectiveness and robustness of the system. 

\subsection{Prediction of Mission in Terms of State-of-Charge}
\label{subsec:prediction}

The prediction of discharge rate is acquired based on distances between mission WPs, the UAV’s velocity, maximum flight time, and initial SoC. The cost of battery replacement is determined by the predicted UAV battery drain during its flight to the battery management station and back to the mission. 

It is necessary to transform each UAV mission WP into a local frame in $x$,\,$y$, or $x$,\,$y$,\,$z$ coordinates in meters. This must be also done for the current location of the UAV and the locations of the battery management stations. Currently, the prediction of the SoC is based on distances between WPs, the current location of the UAV, the distance between WPs and stations, the maximum flight time of the UAV, and the UAV travel speed, which is now considered constant for simplicity.

In the current implementation, there is also the option to designate that the UAV has already flown some WPs from the mission. This allows the user to download the entire mission from the UAV, while simultaneously reflecting its progress.

The distances between each point are evaluated by the L2-norm as
\begin{equation}
    d = \sqrt{\sum_{j=1}^N \left(p_{2,j}-p_{1,j}\right)^2},
\end{equation}
where $d$ is the distance, $p_1$ is the start point, $p_2$ is the end point, and $N$ is the number of elements in vector $p_1$ or $p_2$. The time required for travel between these points, designated as $\Delta t$, is given by
\begin{equation}
    \Delta t = \frac{d}{v_{uav,i}},
\end{equation}
where $v_{uav,i}$ is the travel speed of UAV with ID $i$ in m/s. The SoC consumed by this flight is estimated by
\begin{equation}
    \Delta SoC = \frac{\Delta t}{t_{\max,i}},
\end{equation}
where $t_{\max,i}$ is the maximum flight time of UAV with ID $i$.

\subsection{Procedures}
\label{subsec:procedures}

A plan is sought that optimizes the predicted SoC and evaluated costs for flights to battery management stations. This plan is composed of UAV ID, battery management station ID, and WP index. If the UAV batteries lack sufficient capacity to complete the mission, they are obliged to visit the battery management station.

The task is further limited by a fixed number of charged batteries in the battery management stations. Additionally, the mission planner performs a calculation of the expected flight time to the battery management station. This information is utilized to determine the optimal time slot for the UAVs to enter the schedule, with a margin of safety to avoid potential collisions near the DP platform. 

Before the execution of the A* algorithm, all potential actions from each WP to each station are generated. These actions are accompanied by estimates of their respective SoC and time requirements, which are based on the predictions described in Section \ref{subsec:prediction}. The SoC and time required for the flight to the station are stored in a 3D array, where the indices are the UAV ID, station ID, and WP index. This enables the reuse of these values in the mission planner.

Upon deployment of the mission planner, the plan is transmitted to mission control. Periodically, the plan is sought by updating the parameters based on real-world measurements. Mission control oversees the progress of the mission, transmits commands to the UAVs following the plan, and ensures that battery replacements are conducted at specified intervals.

\subsection{A* Implementation}
\label{subsec:astar}

The vertices of the oriented graph for A* are represented by a vector, where the index corresponds to the UAV's ID and the value corresponds to the number of WPs from the mission that the UAV has already flown. Consequently, the starting vertex is typically set to zero, while the goal vertex is equal to the number of WPs of each mission.

Furthermore, the predicted SoC throughout the mission is stored in the vertex. If these values are examined to ascertain whether they exceed the minimum discharge threshold, the mission can be completed without the necessity of further battery replacement. Moreover, the mission's timeline is stored in the vertex, and it is adjusted in iterations to account for the time required for battery replacement.

The optimal solution is sought based on minimizing the total amount of SoC consumed by each UAV. A* is an algorithm that searches for the solution based on heuristics, which provide an estimate of how close the algorithm is from the current vertex to the solution. In this case, the estimate is the SoC needed to complete all missions from the current vertex.

\begin{figure}
    \centering
    \includegraphics[width=0.8\linewidth]{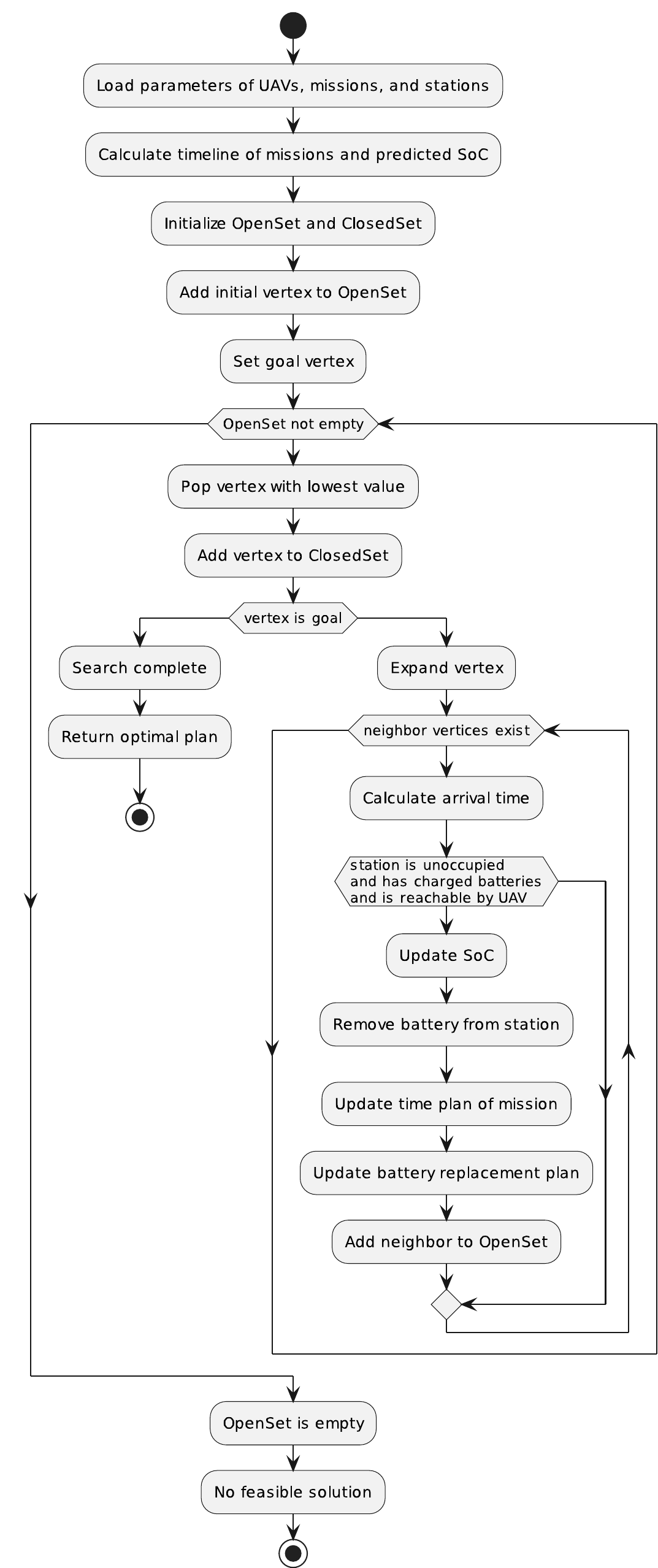}
    \caption{Diagram of the mission planning algorithm for the UAV battery replacement}
    \label{fig:mission-planning}
\end{figure}

For the sake of clarity, a diagram of the mission planning algorithm is presented in Figure \ref{fig:mission-planning}. Once the algorithm has been initialized, it begins an iterative search for a solution, which is the plan of battery replacements that enable the completion of every UAV mission. During this search, the time of arrival is calculated for each potential child of the current vertex.  Additionally, the schedule of the requested station is checked to ensure that it is unoccupied, including safety windows and time for battery replacement.

Subsequently, verification is conducted to ascertain whether the station possesses available charged batteries and whether the UAV is capable of reaching the station without depleting its battery below the critical value specified by the threshold. 

If the action remains viable following these assessments, the newly generated vertex, comprising the updated SoC and the revised time plan for the UAV's mission, is incorporated into the list of potential children on the current vertex.

Once potential children have been identified, it is necessary to ascertain whether they have already been examined. If this is not the case, they are then added to the heap for further investigation, which is a standard procedure in A*.

At the current state of development, the battery object is not replaced between the UAV and the battery station. Instead, the SoC of the battery is simply set to the maximum value. Nevertheless, the occurrence of the battery object provides possibilities for future extensions of the algorithm.

\section{Evaluating Mission Planning Strategies}
\label{sec:results} 

Two distinct scenarios were employed to test the functionality of the mission planner. In both scenarios, several static battery management stations and multiple UAVs were considered, each with its mission. The missions were designed in such a way that completion was not possible for any UAV without changing its battery in the process. In both scenarios, the minimum SoC, which signifies that the UAV is not permitted to fly with a lower value, was set to 20\%.

In the first scenario, the UAV parameters were generated randomly. Their missions were also generated randomly, but to follow the path given by the waypoints on an ellipse. In the second scenario, the parameters of the UAVs were selected according to the Iris UAV, and the mission was planned in the software for planning UAV missions in the form of an area coverage mission above the city park. This mission can be utilized to address real-world use cases where UAVs are, for example, employed in search and rescue operations.

The A* algorithm was implemented in Python based on the implementation described in \cite{Swift2017}, which employs a priority queue represented by a heap.

\subsection{Randomly Generated Missions}
\label{subsec:mission-random}

\begin{figure}
    \centering
    \includegraphics[width=0.8\linewidth]{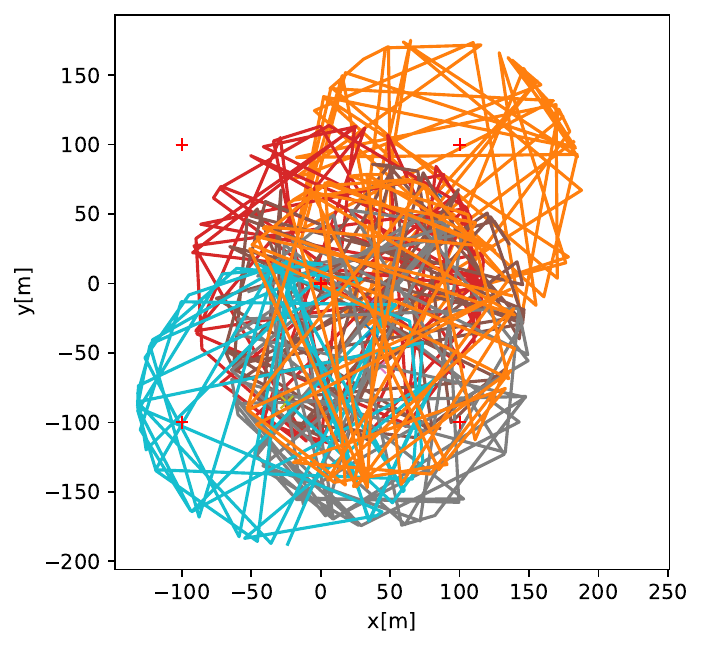}
    \caption{Map with UAV randomly generated missions and battery station scenario}
    \label{fig:map-random}
\end{figure}

Figure \ref{fig:map-random} depicts the generated map, which includes the original UAV missions and the locations of the battery management stations. Each UAV was considered with a different flight speed, initial SoC, and location at the center of the mission ellipse. Each UAV mission consisted of 50 WPs. In the system, 6 UAVs and 5 platforms were considered, each with 10 batteries. All parameters were generated using a uniform distribution. The travel speed ranged from 2 to 6 m/s, with a flight time of 10 to 20 minutes. The initial SoC ranged from 0.6 to 1.0. 

The final solution consisted of 15 battery changes and is presented in Table \ref{tab:actions-random}. Each UAV visited the battery management station 1-3 times. The visits were scheduled according to the distance at the time of the mission and station blockage which schedule is noted in Table \ref{tab:schedule-random}. 

\begin{table}
\centering
\caption{UAV battery replacement actions in randomly generated scenario}
\label{tab:actions-random}
\begin{tabular}{ccc}
\hline
UAV ID & Station ID & WP Index \\
\hline
1 & 204 & 25 \\
1 & 201 & 36 \\
2 & 201 & 18 \\
2 & 203 & 27 \\
2 & 205 & 28 \\
3 & 205 & 21 \\
3 & 204 & 34 \\
3 & 204 & 39 \\
4 & 202 & 23 \\
4 & 201 & 32 \\
5 & 203 & 18 \\
5 & 202 & 29 \\
6 & 202 & 8 \\
6 & 205 & 26 \\
6 & 203 & 35 \\
\hline
\end{tabular}
\end{table}

\begin{table}
\centering
\caption{Schedule of battery stations blocking in randomly generated scenario (min:sec)}
\label{tab:schedule-random}
\begin{tabular}{ccc}
\hline
\text{Station ID} & \text{Start} & \text{End} \\
\hline
201 & 12:11.23 & 15:11.23 \\
201 & 22:01.74 & 25:01.74 \\
201 & 26:52.09 & 29:52.09 \\
202 & 6:31.12 & 9:31.12 \\
202 & 13:04.66 & 16:04.66 \\
202 & 24:30.96 & 27:30.96 \\
203 & 11:57.60 & 14:57.60 \\
203 & 23:27.93 & 26:27.93 \\
203 & 27:40.22 & 30:40.22 \\
204 & 9:07.15 & 12:07.15 \\
204 & 20:01.78 & 23:01.78 \\
204 & 25:48.15 & 28:48.15 \\
205 & 9:50.48 & 12:50.48 \\
205 & 20:13.77 & 23:13.77 \\
205 & 25:56.92 & 28:56.92 \\
\hline
\end{tabular}
\end{table}

\begin{figure}
    \centering
    \includegraphics[width=0.8\linewidth]{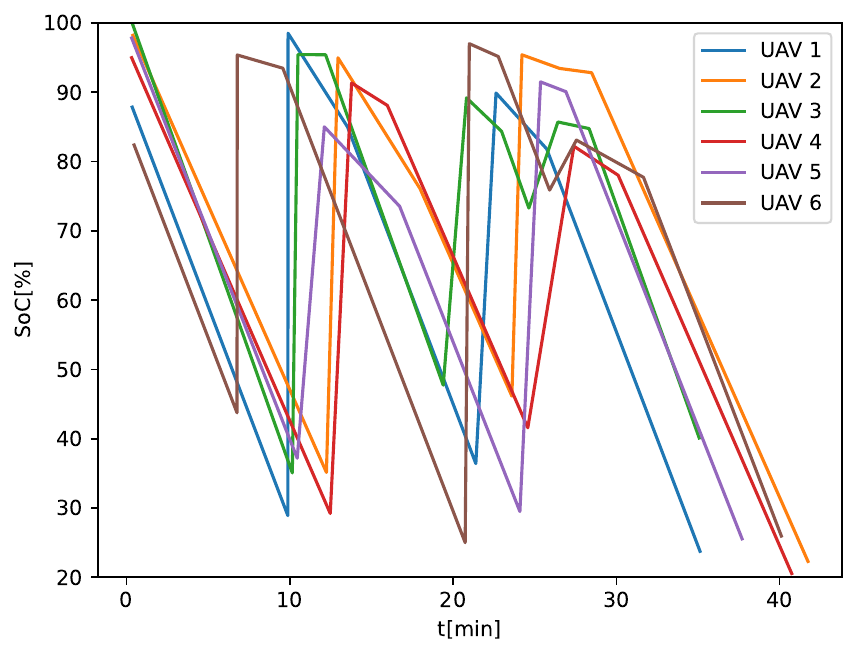}
    \caption{SoC of every UAV in time during randomly generated scenario}
    \label{fig:soc-random}
\end{figure}

The original mission plan without the battery change ranged from 30 to 37 minutes. The final time with the planned battery replacements ranged from 35 to 40 minutes. However, the battery replacement allowed the missions to be accomplished. The SoC level of each UAV during the mission is shown in Figure \ref{fig:soc-random}.

\subsection{Area Coverage Multi-UAV Mission in City Park}
\label{subsec:mission-park}

\begin{figure}
    \centering
    \includegraphics[width=0.8\linewidth]{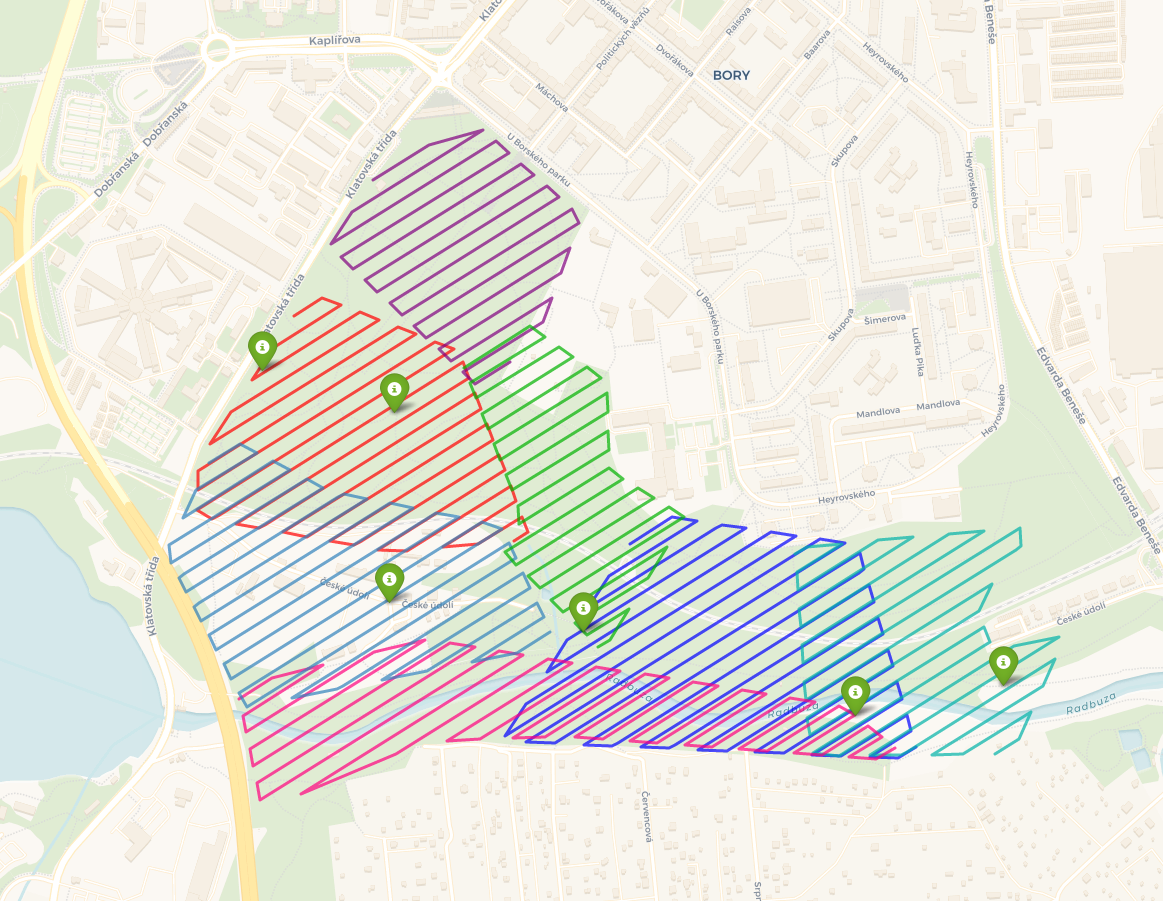}
    \caption{Map with UAV area coverage missions and battery station locations in city park}
    \label{fig:map-park}
\end{figure}

In this scenario, the missions were planned in the existing city park using open-source software for flight control and mission planning, namely QGroundControl (QGC\footnote{QGC -- \url{http://qgroundcontrol.com/}}). Subsequently, the missions were uploaded into the UAV, which was simulated using PX4 Software In The Loop (SITL)\footnote{PX4 SITL -- https://docs.px4.io/main/en/simulation/index.html}, a component of the PX4 Autopilot software. This was followed by a download of the data. A Python implementation of the MAVSDK\footnote{MAVSDK -- \url{https://mavsdk.mavlink.io/main/en/index.html}} library was employed for MAVLink protocol-based communication with UAVs. The data was subsequently saved to a file for further processing. This procedure was designed to emulate the real-world process. Figure \ref{fig:map-park} depicts the map with the missions and the locations of the battery management station. The missions for 7 UAVs included a range from 80 to 130 WPs, and 6 stations were placed in the park.

During the execution of the A* algorithm, we encountered difficulties with the computational time. Even 5 hours proved insufficient to identify the optimal plan. However, the issue was resolved without modifying the A* algorithm by limiting the set of possible actions to those flights to stations that were under 2 minutes from the current WPs. This removed nonviable actions and reduced the time to find a solution to 4 minutes and 37 seconds.

\begin{figure}
    \centering
    \includegraphics[width=0.8\linewidth]{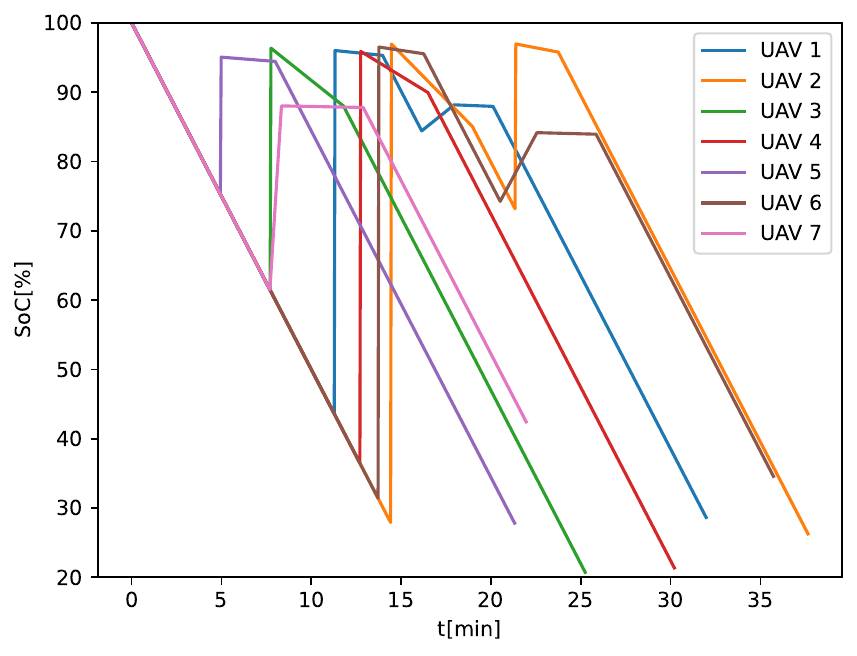}
    \caption{SoC of every UAV in time during area coverage mission in city park}
    \label{fig:soc-park}
\end{figure}

In the city park scenario, 10 battery replacements were planned (see Table \ref{tab:actions-park}. The schedule of replacements is presented in Table \ref{tab:schedule-park}. The original mission plans were designed to last between 17 and 33 minutes but were extended by the flight to stations to last between 22 and 37 minutes. Figure \ref{fig:soc-park} presents the SoC of each UAV. It can be observed that all UAVs started with an SoC of 100\% and the duration of flight varied considerably between them. This variation can be attributed to differences in mission length or the placement of stations.

\begin{table}
\centering
\caption{UAV battery replacement actions in area coverage mission in city park}
\label{tab:actions-park}
\begin{tabular}{ccc}
\hline
UAV ID & Station ID & WP Index \\
\hline
1 & 205 & 31 \\
1 & 201 & 38 \\
2 & 202 & 33 \\
2 & 202 & 39 \\
3 & 203 & 29 \\
4 & 204 & 33 \\
5 & 201 & 15 \\
6 & 201 & 43 \\
6 & 201 & 54 \\
7 & 201 & 38 \\
\hline
\end{tabular}
\end{table}

\begin{table}
\centering
\caption{Schedule of battery stations blocking in area coverage mission in a city park (min:sec)}
\label{tab:schedule-park}
\begin{tabular}{ccc}
\hline
Station ID & Start & End \\
\hline
201 & 4:55.66 & 7:55.66 \\
201 & 9:05.58 & 12:05.58 \\
201 & 13:24.82 & 16:24.82 \\
201 & 17:30.74 & 20:30.74 \\
201 & 22:41.14 & 25:41.14 \\
202 & 14:01.43 & 17:01.43 \\
202 & 20:57.01 & 23:57.01 \\
203 & 7:26.32 & 10:26.32 \\
204 & 12:31.64 & 15:31.64 \\
205 & 11:04.58 & 14:04.58 \\
\hline
\end{tabular}
\end{table}

Each UAV changed its battery once or twice. It is notable that station 201, situated in the center of the red mission in Figure \ref{fig:map-park}, has completed 5 battery replacements. In contrast, station 202 has performed 2 replacements, while station 206 (which is situated second from the east) has not performed any replacements. The remaining stations have completed only a single replacement. This data indicates that the stations are not positioned optimally, suggesting that moving them could potentially lower the requirements for the number of batteries or stations.

\section{Conclusion}
\label{sec:conclusion}

The mission planner for battery management stations was described in detail, including an introduction of its various components. The planning algorithm was tested in two scenarios, one of which placed particular emphasis on its resemblance to the real-world area coverage mission in the existing city park. This was achieved by acquiring the original mission from the UAV simulated with PX4 SITL. In both scenarios, the optimal plan was identified. However, in the area coverage scenario, limitations of the algorithm were encountered due to the large set of possible actions. This issue was successfully addressed by limiting flights from the WPs to the stations to those under 2 minutes. This adjustment of the action set allowed the acquisition of an optimal solution in 4.5 minutes. The mission planner was demonstrated to be functional in these scenarios.

In the future, the mission planner could be enhanced with more realistic UAV trajectories. Furthermore, it would be advantageous to reflect battery charging in battery management stations during the planning process. Additionally, it would be beneficial to implement an enhanced model of battery changes in the SoC. Moreover, the entire system would benefit from the utilization of an algorithm for the online identification of parameters associated with the problem. Furthermore, the algorithm could be modified to enhance the speed of replanning.

\section*{Acknowledgment}

This work was supported by the Technology Agency of the Czech Republic, programme National Competence Centres, project \#TN 0200 0054 Bozek Vehicle Engineering National Competence Center.

\bibliographystyle{IEEEtran}
\bibliography{mybib}

\end{document}